\pdfoutput=1

\documentclass[11pt]{article}

\usepackage[]{acl}

\usepackage{times}
\usepackage{latexsym}

\usepackage[T1]{fontenc}

\usepackage[utf8]{inputenc}

\usepackage{microtype}

\usepackage{inconsolata}
\usepackage{import}
\usepackage{microtype}
\usepackage{layout}
\usepackage{tabularx, makecell}
\usepackage{booktabs}
\usepackage{mathrsfs}
\usepackage{amssymb} 
\usepackage{url}
\usepackage{graphicx}
\usepackage{xspace,paralist}
\usepackage{times,latexsym}
\usepackage{amsmath}
\usepackage{appendix}
\usepackage{comment} 
\usepackage{enumitem}
\usepackage{makecell}
\usepackage{multirow}
\usepackage{xcolor}
\usepackage{graphicx}
\usepackage{cleveref}

\newcommand{\ex}[1]{\textit{#1}\xspace}
\newcommand{\eqnref}[1]{Eq~\eqref{#1}\xspace}
\newcommand{\tabref}[2][]{Table#1~\ref{#2}\xspace}
\newcommand{\figref}[1]{Figure~\ref{#1}\xspace}
\newcommand{\secref}[1]{Section~\ref{#1}\xspace}
\newcommand{\appref}[1]{Appendix~\ref{#1}\xspace}

\newcommand{\dataset}[1]{\text{#1}\xspace}
\newcommand{\stsb}{\dataset{STS-B}}
\newcommand{\ncsts}{\dataset{N2C2-STS}}
\newcommand{\medsts}{\dataset{MedSTS}}
\newcommand{\biosses}{\dataset{BIOSSES}}
\newcommand{\ebmsass}{\dataset{EBMSASS}}
\newcommand{\usts}{\dataset{USTS}}
\newcommand{\ustsu}{\dataset{USTS-U}}
\newcommand{\ustsc}{\dataset{USTS-C}}

\newcommand{\chaosnli}{\dataset{ChaosNLI}}
\newcommand{\chaossnli}{\dataset{Chaos-SNLI}}
\newcommand{\chaosmnli}{\dataset{Chaos-MNLI}}

\newcommand{\mednli}{\dataset{MedNLI}}

\newcommand{\model}[1]{\text{#1}\xspace}

\newcommand{\chainofthought}{\model{CoT}}
\newcommand{\chatgpt}{\model{ChatGPT}}
\newcommand{\gptfour}{\model{GPT-4}}

\newcommand{\llama}{\model{LLaMA}}
\newcommand{\llamatwo}{\model{LLaMA-2}}

\newcommand{\alpaca}{\model{Alpaca}}

\newcommand{\claude}{\model{Claude}}
\newcommand{\bloomz}{\model{BLOOMz}}

\newcommand{\bert}{\model{BERT}}

\usepackage{amssymb}
\usepackage{pifont}
\title{Rethinking STS and NLI in Large Language Models}

\author{Yuxia Wang\textsuperscript{1,3} \quad Minghan Wang\textsuperscript{2} \quad \textbf{Preslav Nakov}\textsuperscript{1} \quad \\
\textsuperscript{1}MBZUAI \qquad \textsuperscript{2}Monash University \qquad \textsuperscript{3}LibrAI\\
  \texttt{\{yuxia.wang, preslav.nakov\}@mbzuai.ac.ae} \\
  \tt{minghan.wang@monash.edu}
}

\begin{document}
\maketitle
\begin{abstract}
Recent years, have seen the rise of large language models (LLMs), where practitioners use task-specific prompts; this was shown to be effective for a variety of tasks. However, when applied to semantic textual similarity (STS) and natural language inference (NLI), the effectiveness of LLMs turns out to be limited by low-resource domain accuracy, model over-confidence, and difficulty to capture the disagreements between human judgements. With this in mind, here we try to rethink STS and NLI in the era of LLMs. We first evaluate the performance of STS and NLI in the clinical/biomedical domain, and then we assess LLMs' predictive confidence and their capability of capturing collective human opinions. We find that these old problems are still to be properly addressed in the era of LLMs.
\end{abstract}

\section{Introduction}
Semantic textual similarity (STS) is a fundamental natural language understanding (NLU) task involving the prediction of the degree of semantic equivalence between two pieces of text~\citep{cer2017semeval}.
Under the regime of first pre-training a language model and then fine-tuning with labelled examples, there are three major challenges in STS modelling (see examples in \tabref{tab:challenges}):
(\emph{i})~low accuracy in low-resource and knowledge-rich domains due to the exposure bias~\citep{wang2020evaluating, wang2020learning};
(\emph{ii})~models make incorrect predictions over-confidently, unreliable estimations are dangerous and may lead to catastrophic errors in safety-critical applications like clinical decision support~\citep{wang2022capture};
(\emph{iii})~difficulty in capturing collective human opinions on individual examples~\citep{wang2022capture}. 
Akin to STS, natural language inference (NLI) faces similar issues, where the goal is to determine whether a \textit{hypothesis} sentence can be entailed from a \textit{premise}, is contradicted, or is neutral with respect to the \textit{premise}.

Large language models (LLMs), such as \chatgpt, \claude and \llamatwo, have demonstrated impressive performance on natural language understanding and reasoning tasks, by simply inputting appropriate prompts or instructions, without any parameter modifications.
On general STS-B~\citep{cer2017semeval}, zero-shot \chatgpt achieves competitive Pearson correlation ($r$) of 80.9 vs. 83.0 by fine-tuning BERT-base using thousands of training examples~\citep{devlin2018bert}.\footnote{Note that \citet{zhong2023can} have reported much higher results of 92.9 using RoBERTa-large on STS-B, but they are calculated on a subset that they sampled from a uniform distribution based on similarity bins, i.e., sampling an equal number of examples binning to 0.0-1.0, 1.0-2.0, 2.0-3.0, 3.0-4.0, and 4.0-5.0, instead of the whole development or test set of STS-B.}
On MNLI-m~\citep{Williams2018mnli}, zero-shot \chatgpt even outperforms fine-tuned RoBERTa-large: accuracy of 89.3 vs. 88.0.\footnote{There might also be data contamination, i.e., the LLM might have seen (part of) the data during training.}
LLMs' remarkable capabilities in zero-shot setting motivate us to rethink the task of STS/NLI and the three challenges under LLM prompt-based generation.

We ask the following questions: (\emph{i})~How well do LLMs perform over knowledge-rich and low-resource domains, such as biomedical and clinical STS/NLI? 
(\emph{ii})~Does the paradigm of prompting LLMs lead to over-confident predictions? 
and (\emph{iii})~How to capture collective human opinion (the distribution of human judgements) using LLMs?

\citet{chen2023howrobust} evaluated GPT-3.5 (\textit{text-davinci-003}) on NLI (e.g.,~SNLI, MNLI, QQP) and on the semantic matching dataset MRPC (it is a binary classification task that predicts whether two sentences are semantically equivalent).
\citet{zhong2023can} evaluated \chatgpt over STS/NLI datasets including STS-B, MNLI, QNLI, and RTE.
We found that they focused on the performance of general-purpose STS and NLI.
However, it is unclear how well \chatgpt performs on clinical and biomedical domains over these two tasks.

\citet{jiang-etal-2021-know} studied the calibration of T5, BART, and GPT-2 on question answering (QA) tasks: whether the model makes well-calibrated predictions, i.e.,~whether the probability it assigns to the outcomes coincides with the frequency with which these outcomes actually occur.
The predictive probability (confidence) will be a reliable signal to assist in deciding how much we can trust a prediction and the corresponding risks we may take.
Unfortunately, the answer is a relatively emphatic \textit{no}.
Most prior work focused on white-box calibration for QA and showed that LLMs are more calibrated on diverse multiple choice QA~\citep{jiang-etal-2021-know, kumar2022answer, kadavath2022language}. 
However, there have been no studies on the calibration of STS/NLI neither in a white-box nor in a black-box scenario. 

Moreover, there are studies exploring LLMs' robustness across NLU tasks, i.e.,~the accuracy variation against adversarial attacks~\citep{chen2023robust}, while less attention has been paid to human disagreement in labelling and how to capture the distribution of multiple individual opinions instead of an aggregated label by averaging or majority voting.
In this work, we aim to bridge these gaps by first evaluating the accuracy of clinical/biomedical STS and NLI over five datasets, and then assessing LLM predictive confidence and their capability of capturing collective human opinions.

We have three major findings:
\begin{itemize}
    \item Fine-tuned BERT-base outperforms zero-shot \chatgpt on nine STS and NLI datasets among ten, involving both general, clinical and biomedical domains, especially on benchmarks where high disagreement exists between individual annotators (\usts and \chaosnli), showing the gap of 0.3 (0.86 vs. 0.56) for Pearson correlation ($r$). \llamatwo (7B, 13B) perform worse despite of using few-shot prompt ($r$=0.58 on \stsb). 
    \item Both black-box and white-box approaches have large calibration error, particularly on STS (continuous label). The larger the LLM, the better calibration: \chatgpt > \llamatwo (13B) > \llamatwo (7B).
    \item LLMs may be able to provide personalised descriptions for a specific topic, or generate semantically-similar content in different tones, but it is hard for current LLMs to make personalised judgements or decisions.   
\end{itemize}
\section{Background}
\label{sec:background}
We first describe STS and NLI, and the datasets we use, and then we discuss three challenges in pretrained language models, followed by how they are approached in LLMs using prompting strategies.

\subsection{Task and Datasets}
\paragraph{Task:}
STS and NLI are both sentence-pair relationship prediction tasks.
STS assesses the degree of semantic equivalence between two snippets of text.
The aim is to predict a continuous similarity score for a sentence pair $(S1, S2)$, generally in the range $[0,5]$, where 0 indicates complete dissimilarity and 5 indicates equivalence in meaning.  
NLI highlights semantic reasoning, determining whether a given \textit{hypothesis} can be logically inferred from a given \textit{premise}, where if it can be, the example falls into \textsc{entailment}), otherwise \textsc{contradiction}, if undetermined \textsc{neutral}.

\paragraph{Datasets:}
For STS, we use two large general datasets --- \stsb~\cite{cer2017semeval} and uncertainty-aware \usts (Chinese) with a collection of annotations for each example~\citep{wang2023collective}, two small clinical datasets --- \medsts~\cite{wang2018medsts} and \ncsts~\cite{wang2020n2c2sts}, and two small biomedical ones --- \biosses~\citep{souganciouglu2017biosses} and \ebmsass~\citep{hassanzadeh2019quantifying}.

For NLI, we use: \mednli, which was annotated by physicians and is grounded in the medical history of patients~\citep{romanov2018mednli}, 
and \chaosnli~\citep{nie2020learn}, which was created by collecting 100 annotations per example for 3,113 examples in SNLI (1,514)~\citep{snli:emnlp2015} and MNLI (1,599)~\citep{Williams2018mnli}, denoted as \chaossnli and \chaosmnli, respectively.
See \appref{sec:appendix-dataset} for statistics of the datasets.

\begin{table}[!t]
\centering
\resizebox{\columnwidth}{!}{
    \begin{tabular}{l p{7cm}}
        \toprule
        \textbf{No.\ 1} & \textsc{Low-resource \& knowledge-rich} \\[1ex]
        S1 & \ex{\underline{Tapentadol} 50 MG Oral tablet 1 tablets by mouth every 4 hours as needed.} \\
        S2 & \ex{\underline{Oxycodone [ROXICODONE]} 5 mg tablet 1 tablets by mouth every 4 hours as needed.} \\
        Gold label & 4.5 \\
        Prediction & 2.0 \\
        Reason & Lack of knowledge: \textit{Tapentadol} and \textit{Oxycodone [ROXICODONE]} are both pain-relief medicine. \\
        \midrule
        \textbf{No.\ 2} & \textsc{Over-confidence wrong prediction} \\[1ex]
        S1 & \ex{You will want to clean the area first.} \\
        S2 & \ex{You will also want to remove the seeds.}  \\
        Gold label & 0.0 \\
        Prediction & $1.95\pm \mathbf{0.004}$ \\
        \midrule
        \textbf{No.\ 3} & \textsc{Capture Human Disagreement} \\[1ex]
        S1 & \ex{A man is carrying a canoe with a dog.} \\
        S2 & \ex{A dog is carrying a man in a canoe.}  \\
        Old label & 1.8 \\
        New label & $\mathcal{N}(\mu=1.7,\sigma=1.0)$ \\[1ex]
        Annotations & [0.0, 0.3, 0.5, 0.5, 1.2, 
                       1.5, 1.5, 1.8, 2.0, 2.0, 
                       2.0, 2.0, 2.5, 3.5, 3.5] \\
        Prediction & 4.3 \\
        Reason & Uncertainty about the impact of key differences in
                 event participants on instances of high lexical overlap \\
        \midrule
        Premise & Look, there's a legend here. \\
        Hypothesis & See, there is a well known hero here. \\
        Old label & (0, 1, 0) \\
        New label & (0.01, 0.57, 0.42) \\
        Annotations & {C: 1, \textbf{E: 57, N: 42}} \\
        Source &  Chaos-MultiNLI \\
        \bottomrule
    \end{tabular}
    }
    \caption{Challenging STS/NLI examples for the PLM-fine-tuned model. ``Old label'' = gold label by averaging or majority voting; ``New label'' = full distribution aggregated over 15 or 100 new ratings;
    and ``Prediction'' = similarity score predicted by fine-tuning the STS model based on BERT-base.}
    \label{tab:challenges}
\end{table}

\subsection{STS/NLI Challenges under PLM}
There are three major challenges in STS and NLI modelling based on the paradigm of fine-tuning a pre-trained language model (PLM) such as BERT~\citep{wang2020learning, wang2022capture, wang2022uncertaintysts, wang2023collective}.

\paragraph{Low accuracy in low-resource domains}
In domains such as biomedical and clinical, domain experts (e.g.,~a physician or a clinician) are required in the annotation process for the data quality, which leads to an extremely-limited amount of labelled data (less than 2,000 examples in clinical/biomedical STS datasets).

Moreover, domain text is rich in specific terms and concepts that rarely appear in a general text.
It is hard for language models that were pre-trained on a general corpus to understand domain terms and the relationship between them due to exposure bias, when the lexical expressions are different.

Example 1 in \tabref{tab:challenges} shows that a clinical STS model tuned on the N2C2-STS training data struggles assigns a semantic similarity score of 2.0 to the sentence pair, while the gold score is 4.5. This is due to the lack of clinical knowledge that \textit{Tapentadol} and \textit{Oxycodone} are both pain-relief medicines.

As current language models have much more capacity and are pre-trained on more data, compared to BERT, do they perform better? How well do LLMs perform on low-resource and knowledge-rich domains? We study this in \secref{sec:domain}.

\paragraph{Over-confidence on wrong predictions}
Neural models have been empirically demonstrated poor calibration --- the predictive probability does not reflect the true correctness likelihood, and they are generally over-confident when they make wrong predictions \cite{guo2017calibration, wang2022uncertaintysts}. 
Put differently, the models do not know what they don't know. 
For No.2 in \tabref{tab:challenges}, the STS model incorrectly predicts the similarity score as 1.95 when the gold label is 0.0.
In such cases, a reliable model should display a high predictive uncertainty (large standard deviation), instead of 0.004.

Faithfully estimating the uncertainty of model predictions is as important as obtaining high accuracy in many safety-critical applications, such as autonomous driving or clinical decision support~\citep{chen2020unite, kendall2017uncertainties}.
If models were able to faithfully reflect their uncertainty when they make erroneous predictions, they could be used reliably in critical decision-making contexts, and avoid catastrophic errors. 
Can LLMs show high confidence when they make correct predictions and low confidence when they make wrong predictions?
How to estimate the predictive confidence/uncertainty in generative LLMs for STS and NLI?
Are the predictions well-calibrated? We will answer these questions in \secref{sec:calibration}.

\paragraph{Capturing collective human opinions}
Due to the task subjectivity and language ambiguity, there exists high disagreement for some cases in STS and NLI labelling, as examples under category No.3 in \tabref{tab:challenges}.
Based on a collection of individual ratings, the average score $\mu$ of {1.7} does not convey the fact that the ratings vary substantially ($\sigma > 1.0$), and the label (0, 1, 0) also does not reflect the inherent disagreements among raters for the NLI example, where there are 57 annotators among 100 assign \textsc{entailment} and 42 assign \textsc{Neutral}.

The gold label aggregated by averaging or majority voting may reflect the average opinion or the majority viewpoint, but fails to capture the latent distribution of human opinions or interpretations, and masks the uncertain nature of subjective assessments.
Simply estimating aggregated labels over examples with high disagreement is close to a random guess of an average opinion.
How to capture the distribution of human opinions under LLMs? Can it be achieved by leveraging LLMs' capability of generating personalised responses under different roles? \secref{sec:humanopinion} offers hints.

\subsection{Are STS/NLI worth studying in LLMs?}
STS and NLI tasks were used to evaluate language models' semantic understanding ability.
LLMs such as \gptfour and \claude have shown remarkable capabilities in following user instructions and helpfully responding a variety of open-domain questions.
This implicitly indicates their great semantic understanding ability.
Moreover, labels of both tasks are sometimes ambiguous and subjective due to the high disagreement between annotators in labelling.
As such, it seems not worthwhile to continue studying STS and NLI anymore under LLMs.

Actually, this is not the whole picture. 
On the one hand, we wonder whether LLMs have the same challenges as PLMs.
On the other hand, we still need accurate and reliable STS/NLI modelling.
STS and NLI focus on analysing semantic relationship between two pieces of text, which allows us to automatically compare, analyse and evaluate LLMs' responses in terms of helpfulness, factuality, bias, toxicity and harmfulness. 
For example, in fact-checking to identify the veracity, STS is the core technique in dense information retrieval to collect the most relevant evidence given a claim, and NLI is always used to identify the stance of the evidence, supporting, refuting or being irrelevant to the claim.
They reduce the human intervention and improve the efficiency.
\begin{table*}[!t]
\centering
\resizebox{\textwidth}{!}{
    \begin{tabular}{l c | c c c | c c r | c c r}
        \toprule
        & \textbf{\bert} & \multicolumn{3}{c|}{\textbf{\chatgpt} Zero-shot} & \multicolumn{3}{c|}{\textbf{\llamatwo (7B)} Few-shot} & \multicolumn{3}{c}{\textbf{\llamatwo (13B)} Few-shot} \\
        \textbf{STS}$\downarrow$ & Base (r) & $r\uparrow$ & $\rho \uparrow$ & MSE $\downarrow$ & $r\uparrow$ & $\rho \uparrow$ & MSE $\downarrow$ & $r\uparrow$ & $\rho \uparrow$ & MSE $\downarrow$ \\
        \midrule
        STS-B & \textbf{0.868} & 0.827 & 0.825 & 1.16 & 0.528 & 0.551 & 3.49 & 0.584 & 0.597 & 2.87\\
        BIOSSES & 0.854 & \textbf{0.865} & 0.888 & 0.56 & 0.181 & 0.129 & 6.73 & 0.254 & 0.223 & 8.50\\
        EBMSASS & \textbf{0.867} & 0.805 & 0.650 & 0.50 & 0.078 & 0.071 & 8.62 & 0.189 & 0.202 & 9.51 \\
        MedSTS & \textbf{0.859} & 0.790 & 0.701 & 0.72 & 0.278 & 0.250 & 2.49 & 0.186 & 0.176 & 3.69 \\
        N2C2-STS & \textbf{0.902} & 0.817 & 0.754 & 0.90 & 0.328 & 0.316 & 6.99 & 0.254 & 0.270 & 9.88\\
        \ustsc (high) & \textbf{0.861} & 0.556 & 0.551 & 2.97 & 0.038 & 0.052 & 11.3 & 0.004 & 0.042 & 10.4 \\
        \ustsu (low) & \textbf{0.838} & 0.552 & 0.465 & 3.09 & 0.076 & 0.096 & 14.6 & 0.107 & 0.129 & 13.1 \\
        \midrule
        \textbf{NLI}$\downarrow$ & Base (Acc) & Acc $\uparrow$ & F1-macro$\uparrow$ & Prec/Recall$\uparrow$  & Acc $\uparrow$ & F1-macro$\uparrow$ & Prec/Recall$\uparrow$  & Acc $\uparrow$ & F1-macro$\uparrow$ & Prec/Recall$\uparrow$  \\
        \midrule
        Chaos-SNLI & \textbf{0.747} & 0.491 & 0.485 & 0.480/0.521 & 0.368 & 0.375 & 0.407/0.452 & 0.350 & 0.319 & 0.314/0.480\\
        Chaos-MNLI & \textbf{0.558} & 0.479 & 0.472 & 0.498/0.509 & 0.348 & 0.306 & 0.361/0.434 & 0.396 & 0.321 & 0.358/0.471 \\
        MedNLI & \textbf{0.777} & 0.739  & 0.743 & 0.763/0.739 & 0.412 & 0.312 & 0.431/0.412 & 0.516 & 0.414 & 0.509/0.516 \\
        \bottomrule
    \end{tabular}
    }
    \caption{Evaluation of zero-shot \chatgpt (helpful assistant) and few-shot \llamatwo (7B, 13B): correlation ($r$, $\rho$) and MSE on seven STS datasets across domains; and precision (Prec), recall and F1 score on three NLI datasets. Baselines (Base) are estimations by fine-tuned STS/NLI model based on \textit{BERT-base}.}
    \label{tab:zeroshot-results}
\end{table*}
\section{Clinical and Biomedical Evaluation}
\label{sec:domain}
How well do LLMs encode clinical and biomedical knowledge, compared with small pretrained language models?

\citet{singhal2023large} assess PaLM (8B to 540B)'s potential in medicine through answering medical questions.
They observed strong performance as a result of scaling and instruction fine-tuning.
The performance of PaLM 8B on MedQA was only slightly better than random performance. Accuracy improved by more than 30\% for PaLM 540B.

\citet{wu2023exploring} evaluate the proprietary LLMs \chatgpt and \gptfour, and open-source models including \llama, \alpaca and \bloomz on a radiology corpus, determining whether a context sentence from a radiology report contains the answer given the question, by the answer of \textit{Yes} or \textit{No}. Results show that \gptfour outperforms \chatgpt, followed by \llama, \alpaca and \bloomz.
Fine-tuning BERT with >1k and >8k task-specific examples can respectively achieve competitive accuracy against 10-shot \chatgpt and 10-shot \gptfour.

We see an ability that does not exist in small models, and rapidly improves above random beyond a certain model size.
How do LLMs perform for clinical and biomedical STS and NLI?

\subsection{Case Study Take-Away} 
Before extensive evaluation, we conduct a case study to investigate what may impact the in-context learning performance for STS and NLI in \appref{sec:casestudy}.
We first study the impact of different prompting strategies: (1) Zero-shot, (2) Zero-shot with annotation guidelines (AG), (3) Zero-shot with chain of thought (CoT), (4) Few-shot, (5) Few-shot with AG, and (6) Few-shot with CoT.

\textbf{How to craft a prompt and parse labels out?}
For prompts with AG, CoT and demonstration exemplars, how will the order of task description, guidelines, CoT and exemplars impact the accuracy? Which order is better? \tabref{tab:prompts} exhibits the final optimised prompts.
Then how to parse the predicted labels out of the free-form responses of LLMs?
We propose to parse the response by model itself when rule-based matching and regular expressions are insufficient, but at the risk of hallucinating a different label.
Experiments show that rule-based parsing obtains better accuracy than model's auto-parsing when the model can follow the instruction and output labels as the requested format.

\textbf{Which prompt performs the best?}
The experiments show that zero-shot performs the best using \chatgpt, and few-shot (w/wt CoT) for \llamatwo.
We speculate that the brief annotation guidelines and limited exemplars may mislead \chatgpt to struggle \textit{what is important information} and \textit{what are unimportant details}, overlooking the overall semantics and failing to make correct judgement.
While for smaller \llamatwo, more information is needed in the context to guide it in track.

\textbf{Why does zero-shot CoT collapse?}
LLMs will give detailed steps of how to calculate a similarity score using different metrics and features when using zero-shot \chainofthought. 
Many responses analyse the similarity score on axes of sentence structure, bag of words, topics and other superficial aspects.
Generally, these score will be summed up and re-scaled to 0-1 or 0-5, sometimes are cut by the maximum range of 5.0 without considering the meaning behind the score. 
Such coarse measurements overlook comparison of the overall semantics, and the incautious re-scaling neglects the meaning behind the score range hurts the accuracy of STS significantly.

\textbf{Impact of the system role and the language of prompt.}
We further investigate: will setting the system role as domain expert or instructing the model to make judgements with specific domain knowledge improve the domain accuracy? The answer is \textit{No}. For models like \chatgpt, it even consistently hurts the performance. This may result from less exposure of such instructions and system roles in tuning stage.
It motivates us to think about, on non-English benchmarks, what language instructions will bring better responses, especially for current LLMs that poorly support non-English languages.
Empirical studies show that English instruction is better on Chinese benchmarks.

\subsection{Experiments}
\textbf{Experimental Setup:}
Based on the findings above, we use zero-shot prompt for \chatgpt, few-shot for \llamatwo, and English prompts for Chinese \ustsc and \ustsu.
Ten general, clinical and biomedical STS/NLI datasets are involved.
\ustsc, \chaossnli, and \chaosmnli are composed of ambiguous cases in which high human disagreement exists among annotators.

\textbf{Baselines:}
We reproduce the baseline results from \citet{wang2020evaluating, wang2020learning, wang2022capture, wang2022uncertaintysts, wang2023collective}.
STS-B, MedSTS, N2C2-STS, \ustsc and \ustsu are predicted by \textit{BERT-base} fine-tuned over the training data of corresponding dataset, coupled with data augmentation strategies. For datasets without training data, BIOSSES uses the fine-tuned N2C2-STS model and EBMSASS uses fine-tuned STS-B.
Chaos-SNLI/MNLI are predicted by \textit{BERT-base} fine-tuned over combination of SNLI and MNLI training data, and MedNLI uses fined-tuned \bert by MedNLI training data.

\textbf{Results:}
Estimations by \chatgpt are inferior to baseline predictions of the fine-tuned \textit{BERT-base}, except for comparable results on BIOSSES. \llamatwo performs much worse than \chatgpt, though 13B is better than 7B, where the best $r$ is 0.58 on the general \stsb using 13B model.
This suggests that clinical and biomedical domains remain challenging for a LLM even if it is as powerful as \chatgpt, putting aside open-source smaller-size language models.
Pearson correlation of 0.55 on \ustsc, \ustsu and less than 50\% accuracy on \chaossnli and \chaosmnli reveal that Chinese STS sentence pairs and NLI cases with controversial labels are particularly hard to predict correctly, even for \chatgpt.
\llamatwo collapses on the two Chinese test sets ($r$ is close to 0), showing poor capability of non-English languages.

\section{Calibration under LLM}
\label{sec:calibration}
Calibration measures how well the predictive confidence aligns with the real correctness likelihood.
Depending on a well-calibrated model, we can trust how certain a model is for a correct prediction, and then deliver tasks to human experts when the model is highly uncertain.

\subsection{Challenges}
Differences between textual discriminative and generative models pose challenges in LLM calibration for accuracy calculation and confidence estimation.

\paragraph{Accuracy Calculation:} 
Accuracy can be easily calculated in the classification task where the decision space is clearly defined among the given classes.
However, the distribution of casual generation from large language models is complicated and intricate.

It is ambiguous to scope the label space, given that the golden semantics can be expressed in various ways~\citep{kuhn2023semantic}.
For STS and NLI, we alleviate this issue by prompting LLMs with task-specific instructions that constrain label space, so that generated text contains predicted labels.

\paragraph{Confidence Estimation:}
For a classifier, the probabilistic outputs from \textit{softmax} with logits passing through often serve as the predictive confidence.
For continuous labels, predictive uncertainty is practically represented by standard deviation~\citep{wang2022uncertaintysts}.
However, how to estimate predictive confidence for STS and NLI under generative models is an open question, particularly for black-box LLMs such as \chatgpt, we can only access to the generated text by APIs, without the predictive probability of the next token.

\subsection{Predictive Confidence Estimation}
A good confidence estimation is expected to truly reflect a model's uncertainty in predicting or making decisions. 
We elaborate our approaches to estimating predictive confidence for LLMs, in both black-box and white-box settings below.

\textbf{Black-box LLMs:}
We generate $K$ samples given an example, and then calculate the mean and the standard deviation for STS and the empirical probability for NLI, similarly to \citet{lin2023generate, kuhn2023semantic}, but we skip their step of incorporating the similarity between any two samples, since we parse the label out of free-form responses.

\textbf{White-box LLMs:}
We aim to use the vocabulary probability of the first newly-generated token as the predictive confidence.
This requires a prompt that can generate an output, in which the first token could appear in the label space of STS or NLI in a high probability.
To achieve this, we use few-shot prompts to demonstrate and constrain the output format of the model, guiding the model to sample the first token aligned with the label space.

Practically, after obtaining the output logits from the last token of the prompt, we normalise it into a probability distribution by \textit{softmax}.
For STS with a continuous label space ranging from 0.0 to 5.0, we simplify the experiments by only studying the probability of the integer part, corresponding to the tokens \texttt{[0,1,2,3,4,5]}.
For NLI, we show cases and instruct the model to output lowercase labels, so that it can fall into the three sub-words: \texttt{[\_ent, \_neutral, \_contradiction]}, meeting the probability for entailment, neutral and contradiction.

To examine whether the model can follow the instruction and output the predicted label in the first token, we count how many percentage of examples where the highest probability token is in the label space; and the top3-probable tokens contain label-space tokens (see \tabref{tab:topprobtokens} in \appref{sec:whiteboxlabelprob}).
Almost 100\% examples follow the instruction, generating a label-space token in the first token at a high probability of $\ge$0.8 based on \llamatwo (7B). 
This suggests that proper prompts can lead model to generate labels, effectively supporting white-box predictive confidence estimation.

\begin{table}[!t]
\centering
\resizebox{\columnwidth}{!}{
    \begin{tabular}{l | c c c | c c c | c c c }
        \toprule
        Model$\rightarrow$ & \multicolumn{3}{c|}{\textbf{\chatgpt}} & \multicolumn{3}{c|}{\textbf{\llamatwo (7B)}} & \multicolumn{3}{c}{\textbf{\llamatwo (13B)}} \\
        Dataset$\downarrow$ & $r\uparrow$ & F1$\uparrow$ & ECE$\downarrow$ 
        & $r\uparrow$ & F1$\uparrow$ & ECE$\downarrow$ 
        & $r\uparrow$ & F1$\uparrow$ & ECE$\downarrow$ \\
        \midrule
        \medsts & 0.801 & -- &  \textit{0.622} & 0.269 & 0.076 & 0.818 & 0.252 & 0.087 & 0.754 \\
        \biosses & 0.849 & -- & \textit{1.096} & 0.107 & 0.017 & 0.840 & 0.272 & 0.010 & 0.723\\
        \ustsc & 0.809 & -- & \textit{1.442} & -0.268 & 0.007 & 0.751 & -0.102 & 0.023 & 0.664 \\
        \midrule
        \mednli & -- & 0.668 & 0.238 & -- & 0.312 & 0.457 & -- & 0.407 & 0.277\\
        \chaosnli & -- & 0.541 & 0.215 & -- & 0.356 & 0.418 & -- & 0.309 & 0.348\\
        \bottomrule
    \end{tabular}
    }
    \caption{Pearson correlation ($r$), F1 and ECE for STS/NLI by \chatgpt and \llamatwo (7B, 13B). Note that calculation formula of ECE for STS under \chatgpt is different from others (\textit{italic numbers}), they cannot be compared directly.}
    \label{tab:ece}
\end{table}

\subsection{Experiments}
\paragraph{Metrics}
Expected calibration error (ECE) is applied to measure if the predictive confidence estimates are aligned with the empirical correctness likelihoods.
The perfectly-calibrated model has ECE=0. The lower ECE, the better calibrated.
For STS in black-box setting, we calculate ECE using the formula for continuous values with the mean and standard deviation as \citet{wang2022uncertaintysts},\footnote{By this formula, ECE>1.0 indicates very poor calibration.} while for NLI and white-box STS, we use \eqnref{eq:ece}:
\begin{equation}
    \small
    \sum_{m=1}^M \frac{\lvert B_m \rvert}{n}  \lvert acc(B_m) - conf(B_m)\rvert)
    \label{eq:ece}
\end{equation}

\paragraph{Experimental Setup}
Based on \medsts, \biosses, \ustsc for STS, and \mednli, \chaosnli for NLI,\footnote{We use 200 samples for \ustsc and \chaosnli, same subset as \secref{sec:humanopinion}} we experiment with \chatgpt as the black-box and \llamatwo (7B, 13B) as the white-box proxy.
In a black-box setting, we sample $K$ times ($K$=10 with a zero-shot prompt), and we use standard deviation for continuous labels and the probability for each class for classification outputs as a confidence score.
In a white-box setting, we use the length-normalised joint probability for both STS and NLI.

\paragraph{Results and Analysis}
\chatgpt achieves the lowest calibration error, and also much higher correlation and F1 across all datasets than \llamatwo, as shown in \tabref{tab:ece}.
13B is more calibrated than 7B thanks to being less confident.
\llamatwo exhibits lower ECE and higher F1 in NLI task than the STS.
Large ECE (>0.8) using 7B on STS should be attributed to the large gap between low accuracy (0.22, 0.05 and 0.005) and high confidence (0.82, 0.84 and 0.75 in \tabref{tab:topprobtokens}).
Under satisfying correlation for STS by \chatgpt, it still offers large ECE.
This indicates that over-confidence remains a challenge in LLMs for STS and NLI tasks.




\section{Collective Human Opinion}
\label{sec:humanopinion}
Capturing the distribution of human opinions under large neural models is non-trivial, especially for continuous values. 
Applying Bayesian estimation to all model parameters in large language models is theoretically possible, in practice it is prohibitively expensive in both model training and evaluation.
Deriving uncertainty estimates by integrating over millions of model parameters, and initialising the
prior distribution for each are both non-trivial~\citep{wang2022uncertaintysts}.

Bypassing estimating key parameters of a standard distribution (e.g.\ $\mu$ and $\sigma$ in a Gaussian distribution) to fit the collective human opinions, in this work, we propose estimating personalised ratings which simulate individual annotations, and then compare the two collective distributions.
Specifically, we prompt LLMs by setting the system role with different personas characterised by age, gender, educational background, profession and other skills.
It is assumed that LLMs can make persona-specific judgement within the capability and background of the role.

\paragraph{Hypothesis:}
If language models are capable to do personalised assignments that match the ability of different roles, a helpful assistant should give more accurate estimations than a five-year old child on the complex semantic reasoning tasks, and a linguistic expert is better than an assistant, a NLP PhD student should have comparable judgement to a NLP expert.
Judgements collected from different roles should be close to the distribution of the collective human opinions gathered by crowd-sourcing.

\subsection{Experiment Setup}
Given an example in \chaosnli for NLI and \ustsc for STS, multiple annotations are available to represent the collective human opinions.
We randomly sampled 200 examples from \ustsc, with a similarity score uniformly spanning across 0-5.
We sample 100 cases from \chaossnli and 100 from \chaosmnli, resulting in \chaosnli (200),  to investigate whether \chatgpt can imitate individual ratings under different roles.

\begin{table}[!t]
\centering
\resizebox{\columnwidth}{!}{
    \begin{tabular}{l | c c c c | c c c}
        \toprule
        Dataset$\rightarrow$  & \multicolumn{4}{c|}{\chaosnli} & \multicolumn{3}{c}{\ustsc} \\
        \textbf{System role} $\downarrow$ & Acc$\uparrow$ & Prec$\uparrow$ & Recall$\uparrow$ & F1-macro$\uparrow$ & $r\uparrow$ & $\rho \uparrow$ & MSE $\downarrow$ \\
        \midrule
        Helpful assistant (HA) & 0.525 & 0.504 & 0.522 & 0.506 & 0.656 & 0.684 & 3.32 \\
        HA good at semantic reasoning & 0.475 & 0.491 & 0.480 & 0.463 & 0.702 & 0.727 & 2.78 \\
        HA good at NLI & 0.535 & 0.512 & 0.516 & 0.509 & 0.644 & 0.675 & 2.97 \\
        NLP expert & 0.530 & 0.527 & 0.524 & 0.511 & 0.679 & 0.736 & 3.20 \\
        NLP PhD student & \textbf{0.565} & \textbf{0.557} & \textbf{0.563} & \textbf{0.548} & 0.685 & 0.703 & 3.04 \\
        Data annotator & \textbf{0.565} & 0.533 & 0.543 & 0.534 & 0.639 & 0.696 & 3.57 \\
        Linguistic expert & 0.485 & 0.480 & 0.488 & 0.469 & \textbf{0.758} & \textbf{0.796} & \textbf{2.73} \\
        Google senior engineer & 0.520 & 0.487 & 0.496 & 0.489 & 0.654 & 0.700 & 3.62 \\
        Professional data scientist & 0.510 & 0.493 & 0.504 & 0.490 & 0.667 & 0.728 & 3.50 \\
        Five-year old child & 0.505 & 0.491 & 0.519 & 0.492 & 0.659 & 0.685 & 2.86 \\
        \midrule
        Ensemble & 0.560 & 0.538 & 0.544 & 0.533 & 0.786 & 0.813 & 2.83 \\
        \bottomrule
    \end{tabular}
    }
    \caption{\chaosnli and \ustsc performance under ten different system roles against the aggregated labels of collective human opinions. Aggregation: majority voting for NLI and averaging for STS. Ensemble refers to aggregating predictions of ten roles.}
    \label{tab:humancollectiveopinion}
\end{table}


\subsection{Results and Analysis}
\textbf{Performance differs under different roles.}
However, the model uncertainty may contribute more to the judgement divergence, instead of the personalised opinion.
On samples of \chaosnli and \ustsc, the accuracy differs significantly under different system roles. NLP PhD student performs the best on \chaosnli and the linguistic expert is the best on \ustsc. 
However, how is the distinction affected by the setup of different roles in the pre-context versus the model predictive uncertainty?
If the deviation of multiple runs under the same role is notably smaller than the variance stemming from various roles setting, and a relatively-high performance consistently appears in the well-performed role, we believe that the model is capable to make persona-specific judgement under different roles.
In other words, the setting of different roles in the pre-context may unlock multiple reasoning paths, an optimal role leads reasoning route to more correct answers.

\begin{figure}[t!]
	\centering
        \includegraphics[scale=0.45]{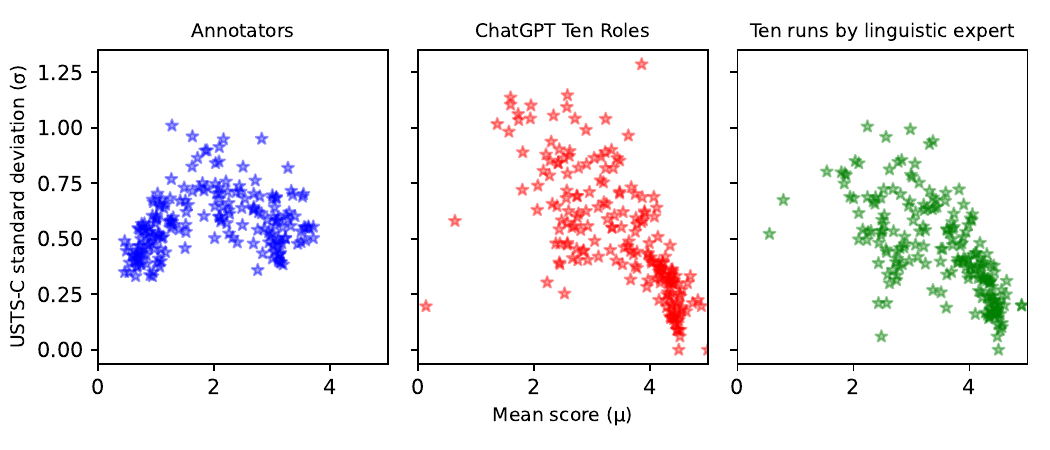}
        \includegraphics[scale=0.45]{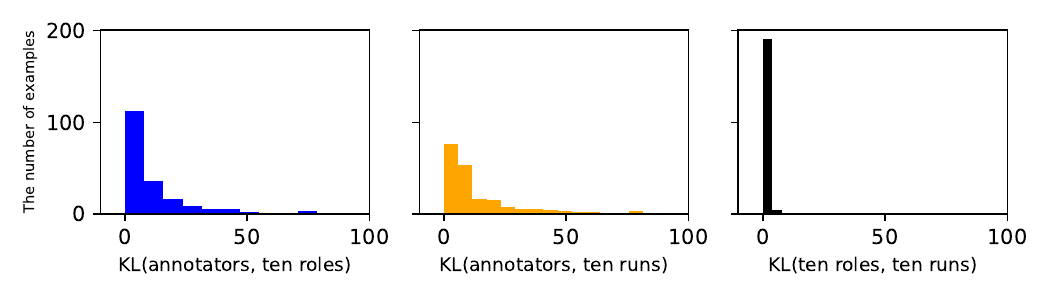}
	\caption{\ustsc ($\mu$, $\sigma$) distribution of annotators versus \chatgpt roles and ten runs by the role of \textit{linguistic expert}, and KL-Divergence (bottom) between the collective human opinions and the distribution of predictions by ten different roles using \chatgpt.}
	\label{fig:ustsc-systemrole-divergence}
\end{figure}

Therefore, we re-run ten times on \chaosnli and \ustsc with the roles of an NLP PhD student and a linguistic expert, respectively.
We cab see in \tabref{tab:tenruns} that, on both \chaosnli and \ustsc, the results deviate significantly across the ten runs.
A higher performance cannot be kept.

The accuracy of \chaosnli ranges from 0.48 to 0.55, and Pearson correlation for \ustsc also ranges from 0.67 to 0.76.
This suggests that the model uncertainty may contribute more to the performance variance, than the setting of system roles.

\textbf{The collective predictions essentially does not match the human opinions.}
Label distributions represented by ($\mu$, $\sigma$) of \ustsc annotators and predictions of ten different roles differ substantially (see \figref{fig:ustsc-systemrole-divergence} top).
The distribution by ten roles and ten runs by \textit{linguistic expert} is much similar, their KL-divergence of 171 (86\%) examples is less than 1.0, indicating small distributional distance for the majority cases between using the same role and different roles. 
While KL-divergence between annotators and ten roles or ten runs is mostly large (KL>1.0 for 177 and 185 examples).
This suggests that neither estimations under different roles nor multiple runs by the same role can imitate the distribution of collective human opinions.

\begin{figure}[t!]
	\centering
        \includegraphics[scale=0.45]{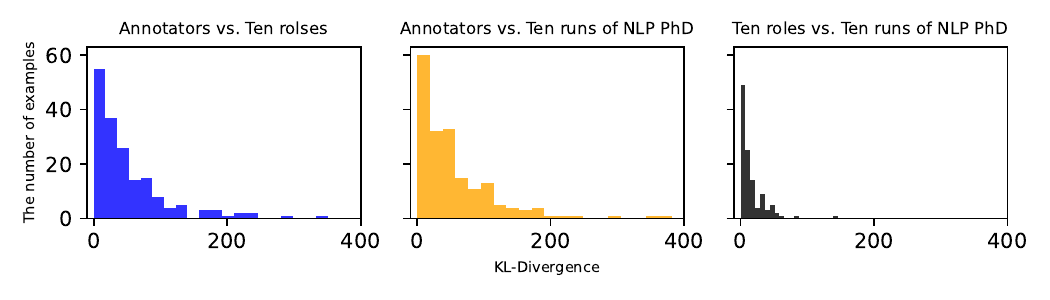}
        \includegraphics[scale=0.45]{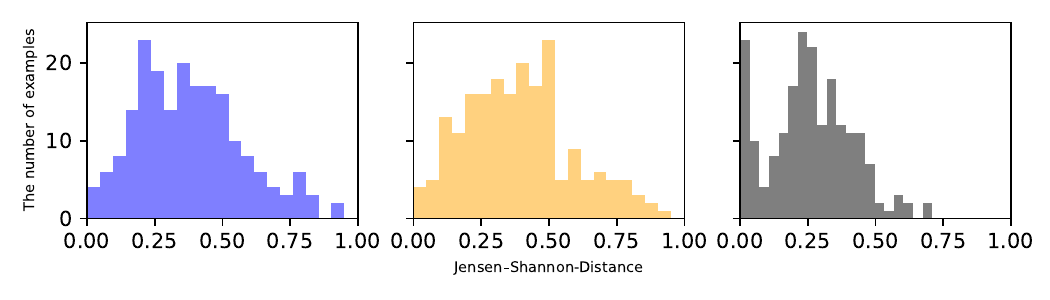}
	\caption{\chaosnli KL-Divergence (top) and Jensen–Shannon distance (bottom) between the collective human opinions and the distribution with bootstrap under predictions by ten different roles using \chatgpt. KL highly correlates with JSD ($r \ge$0.88 and $\rho \ge$ 0.97).}
	\label{fig:chaosnli-systemrole-divergence}
\end{figure}

Similarly, in \figref{fig:chaosnli-systemrole-divergence} for \chaosnli, the distributional divergence between annotators and simulated raters (system roles) spans from 0 to 400, while KL-divergence between ten roles and ten runs in the same role is much smaller, with the majority concentrating within 50.\footnote{Bootstrap is applied to sample 100 judgements, imitating 100 annotations in \chaosnli.}
Moreover, distributions of both KL and JSD of (annotators, ten roles) and (annotators, ten runs under the role of PhD student) are similar.
It indicates that the impact of setting different roles is similar to running multiple times under the same role.

We can conclude that prompting using different roles cannot unlock the LLM's capability of making personalised judgement.

\section{Conclusion and Future Work}
In this study, we aim to rethink STS and NLI challenges in the context of LLMs, to identify whether LLMs alleviate the three issues in the era of \bert.

Experiments on ten STS/NLI datasets show that fine-tuned BERT-base outperforms zero-shot \chatgpt, especially on non-English corpus and ambiguous examples where high disagreement exists between individual annotations. 
Smaller LLMs such as \llamatwo (7B, 13B) collapse if only by in-context learning.
Though the larger model shows smaller calibration error, LLM \chatgpt is still far from a well-calibrated model. 
LLMs may be able to provide personalised descriptions for a specific topic, or to generate semantically similar content in different tones, but it is still hard for current LLMs to make personalised judgements.
These reveal that old problems are not addressed in the new era.

\section*{Limitations}

\paragraph{Prompt optimisation}
Prompt engineering is often important for LLMs to achieve good performance. In this study, we designed and refined prompts for STS and NLI tasks manually.
Though we made efforts to optimise, it is challenging for authors to search the optimal prompt in the large and discrete prompt space.
The inferior prompts may lock the real capabilities of LLMs.
Automatic prompt optimisation algorithm like \citet{deepmind2023promptoptimizer} will be used to customise task-specific and model-specific prompts in our future work. 

\paragraph{More Tasks and More LLMs}
We only evaluate STS and NLI tasks over five biomedical and clinical datasets, this would be insufficient to truly evaluate LLMs' capability in biomedical and clinical domains.
More reasoning-intensive tasks such as questions answering and entity linking can be incorporated.
Moreover, larger open-source language models (e.g., \llamatwo 70B) should be assessed.

\paragraph{White-box Confidence Estimation}
To simplify the confidence estimation in white-box setting, we use probabilities of the label-space tokens. This could be optimised further, particularly for scalar labels in STS.

\section*{Ethics Statement}
This paper respects existing intellectual property by making use of only publicly and freely available datasets.

\paragraph{Biases:}
The study randomly samples ten roles that are either commonly used in research papers or the roles with which authors are familiar, to simulate collective human distributions of STS judgement. It does not consider the real demographic distribution, possibly resulting in some biases. Given that it is just an exploratory case study, less serious harms will be caused.

\paragraph{Healthcare Concern:}
This research investigates the capability of LLMs in biomedical and clinical domains over STS and NLI tasks.
They might be combined to a tool that can be used by healthcare providers, administrators, and consumers, which will require significant additional research to ensure the safety, reliability, efficacy, and privacy of the technology. Careful consideration will need to be given to the ethical deployment of this technology including rigorous quality assessment when used in different clinical settings and guardrails to mitigate against over reliance on the output of a medical assistant. 


\bibliography{ref}
\bibliographystyle{acl_natbib}

\clearpage

\section*{Appendix}
\appendix
\section{Statistics about the Datasets}
\label{sec:appendix-dataset}
\tabref{tab:datasets} shows the statistic information for all datasets used in this paper.
\begin{table*}[ht]
\centering
    \begin{tabular}{l l l l l c l}
        \toprule
        Dataset & \#Train & \#Dev & \#Test & Range & \#Annotation & Domain \\
        \midrule
        STS-B (2017) & 5,749 & 1,500 & 1,379 & $[0,5]$ & 5 & general \\
        MedSTS (2018) & 750 & --- & 318      & $[0,5]$ & 2 & clinical \\
        N2C2-STS (2019) & 1642 & --- & 412   & $[0,5]$ & 2 & clinical \\
        BIOSSES (2017) & --- & --- & 100     & $[0,4]$ & 5 & biomedical \\
        EBMSASS (2019) & --- & --- & 1,000   & $[1,5]$ & 5 & biomedical \\
        \midrule
        USTS-U (2023) & 4,900 & 2,000 & 2,000 & $[0,5]$ & 4 & general \\
        USTS-C (2023) & 2,051 & 2,000 & 2,000 & $[0,5]$ & 19 & general \\
        \midrule
        MedNLI & 11,232 & 1,395 & 1,422  & 3-class & -- & clinical \\
        \midrule
        Chaos-SNLI (2020) & --- & --- & 1,514 & 3-class & 100 & general \\
        Chaos-MNLI (2020) & --- & --- & 1,599 & 3-class & 100 & general \\
        \bottomrule
    \end{tabular}
    \caption{STS/NLI datasets. \#Train, Dev, Test Size = number of text pairs, range = label range. \#Annotator = number of raw annotations for each example.}
    \label{tab:datasets}
\end{table*}

\begin{table}[ht!]
\centering
\resizebox{\columnwidth}{!}{
    \begin{tabular}{c p{7cm}}
        \toprule
        \textbf{Task} & \textbf{Prompt Template} \\
        \midrule
        STS & \textsc{Zero-shot} \\[1ex]
        &     Determine the similarity between the following two sentences (S1, S2). The score should be ranging from 0.0 to 5.0, and can be a decimal. S1: \{\} S2: \{\} Score: \\
        \midrule
        STS & \textsc{Zero-shot (AG)} \\[1ex]
        &     Annotation instructions + Task description. \\
        &     S1: \{\} S2: \{\} Score: \\
        \midrule
        STS & \textsc{Zero-shot (CoT)} \\[1ex]
        &     Determine the similarity between the following two sentences (S1, S2). \underline{\textit{Explain the assessment step by step}}. The score should be ranging from 0.0 to 5.0, and can be a decimal. \\
        &     S1: \{\} S2: \{\} Score: \\
        \midrule
        STS & \textsc{Few-shot} \\[1ex]
        &     Five demonstration examples $\cdots$ \\
        &     Task description. S1: \{\} S2: \{\} Score: \\
        \midrule
        STS & \textsc{Few-shot (AG)} \\[1ex]
        &     Annotation instructions + Five demonstrations + Task description. S1: \{\} S2: \{\} Score: \\
        \midrule
        STS & \textsc{Few-shot (CoT)} \\[1ex]
        &     Task description + Five demonstrations with explanation for each, e.g.,\\
        &     S1: A woman is washing her hands. \\ 
        &     S2: A woman is straightening her hair. \\
        &     Explain: S1 and S2 are in the same topic, but important information is totally different. \\
        &     Score: 0.8 \\ [1ex]
        &     S1: \{\} S2: \{\} \\
        \midrule
        NLI & \textsc{Zero-shot} \\[1ex]
        &     Given the sentence \{\}, determine if the following statement is entailed or contradicted or neutral: \{\}. \\
        \midrule
        NLI & \textsc{Few-shot} \\[1ex]
        &     Given the premise sentence S1, determine if the hypothesis sentence S2 is entailed or contradicted or neutral, by three labels: entailment, contradiction, neutral.\\
        & Six demonstrations (two for each label) \\
        & S1: \{\} S2: \{\} Label: \\
        \bottomrule
    \end{tabular}
    }
    \caption{Summary of the prompt templates we used for the STS and the NLI tasks in the zero-shot and the few-shot prompt settings. CoT stands for chain of thought, and AG stands for annotation guidelines. The task description is the same as for the zero-shot prompt setting.}
    \label{tab:prompts}
\end{table}

\begin{table*}[ht]
\centering
\resizebox{\textwidth}{!}{
    \begin{tabular}{l | c c c c | c c c c | c c c c | c c c c}
        \toprule
        \textbf{Model} $\rightarrow$ & \multicolumn{8}{c|}{\textbf{\chatgpt}} & \multicolumn{8}{c}{\textbf{\llamatwo (7B)}} \\
        Dataset $\rightarrow$ & \multicolumn{4}{c|}{\textbf{\stsb}} & \multicolumn{4}{c|}{\textbf{\ncsts}} & \multicolumn{4}{c|}{\textbf{\stsb}} & \multicolumn{4}{c}{\textbf{\ncsts}} \\
        Prompt Strategy $\downarrow$ & \#valid & $r\uparrow$ & $\rho \uparrow$ & MSE $\downarrow$ & \#valid & $r\uparrow$ & $\rho \uparrow$ & MSE $\downarrow$ &
        \#valid & $r\uparrow$ & $\rho \uparrow$ & MSE $\downarrow$ & \#valid & $r\uparrow$ & $\rho \uparrow$ & MSE $\downarrow$ \\
        \midrule
        zero-shot & 1379 & 0.758 & 0.766 & 1.87 & 412 & \textbf{0.817} & \textbf{0.754} & \textbf{0.90}  
        & 1292 & 0.044 & 0.106 & 4.56 & 378 & -0.065 & -0.013 & 5.93 \\
        
        zero-shot (AG) & 1379 & 0.640 & 0.638 & 1.59 & 412 & 0.532 & 0.531 & 2.53 
        & 1356 & 0.375 & 0.314 & \textbf{2.24} & 402 & 0.228 & 0.196 & \textbf{3.73} \\
        zero-shot (CoT) & 1379 & 0.019 & 0.054 & 489 & 368 & 0.173 & 0.185 & 3.75 
        & 1147 & 0.147 & 0.158 & 4.27 & 388 & 0.018 & 0.012 & 4.99 \\
        \midrule
        few-shot & 1324 & 0.688 & 0.75 & 2.14 & 393 & 0.533 & 0.514 & 3.49 
        & 1373 & 0.506 & 0.423 & 3.26 & 407 & \textbf{0.327} & \textbf{0.317} & 6.97 \\
        few-shot (AG) & 1377 & 0.700 & 0.756 & 1.79 & 389 & 0.505 & 0.469 & 3.03
         & 1375 & 0.436 & 0.383 & 4.06 & 405 & 0.266 & 0.244 & 6.87 \\
        few-shot (CoT) & 1316 & \textbf{0.796} & \textbf{0.796} & \textbf{1.56} & 412 & 0.637 & 0.680 & 3.18 
        & 1351 & \textbf{0.668} & \textbf{0.658} & 2.60 & 397 & -0.029 & -0.183 & 11.02 \\
        \bottomrule
    \end{tabular}
    }
    \caption{\textbf{Impact of prompt strategy:} Pearson ($r$), Spearman ($\rho$) correlation and MSE of general \stsb (1379) and clinical \ncsts (412) test sets using six different prompt strategies: AG = annotation guidelines, CoT = chain of thought. \#valid = the number of valid predictions, where the invalid cases are either refused to respond by LLMs or hard to parse the similarity score from the free-form text by simple rules and LLM auto-parsing.}
    \label{tab:promptimpact}
\end{table*}

\begin{table*}[!t]
\centering
\resizebox{\textwidth}{!}{%
\begin{tabular}{@{}l|cccc|cccc|cccc|cccc|cccc@{}}
\toprule
\textbf{Dataset $\rightarrow$} &
  \multicolumn{4}{c|}{\textbf{\medsts}} &
  \multicolumn{4}{c|}{\textbf{\biosses}} &
  \multicolumn{4}{c|}{\textbf{\ebmsass}} &
  \multicolumn{4}{c|}{\textbf{\ustsc}} &
  \multicolumn{4}{c}{\textbf{\ustsu}} \\ \midrule
Prompt Strategy $\downarrow$ &
  \#valid &
  $r\uparrow$ &
  $\rho \uparrow$ &
  MSE $\downarrow$ &
  \#valid &
  $r\uparrow$ &
  $\rho \uparrow$ &
  MSE $\downarrow$ &
  \#valid &
  $r\uparrow$ &
  $\rho \uparrow$ &
  MSE $\downarrow$ &
  \#valid &
  $r\uparrow$ &
  $\rho \uparrow$ &
  MSE $\downarrow$ &
  \#valid &
  $r\uparrow$ &
  $\rho \uparrow$ &
  MSE $\downarrow$ \\ \midrule
zero-shot       & 297 & 0.007 & 0.036 & 4.83 & 93  & 0.215  & 0.217  & 3.39 & 927 & 0.093 & 0.122  & 3.98  & 1893 & -0.016 & 0.017  & 4.71  & 1896 & 0.029 & 0.096 & 6.05  \\
zero-shot (AG)  & 308 & 0.032 & 0.060 & 1.86 & 97  & 0.109  & 0.116  & 3.00 & 969 & 0.090 & 0.108  & 3.31  & 1994 & 0.040  & 0.039  & 4.69  & 1990 & 0.045 & 0.010 & 6.91  \\
zero-shot (CoT) & 300 & 0.051 & 0.069 & 2.83 & 98  & -0.173 & -0.078 & 4.03 & 972 & 0.048 & 0.071  & 4.01  & 1781 & -0.008 & -0.008 & 4.16  & 1789 & 0.050 & 0.050 & 5.89  \\ \midrule
few-shot        & 305 & 0.255 & 0.272 & 2.48 & 98  & 0.151  & 0.107  & 6.78 & 991 & 0.081 & 0.072  & 8.59  & 1985 & 0.033  & 0.051  & 11.25 & 1993 & 0.076 & 0.091 & 14.58 \\
few-shot (AG)   & 312 & 0.200 & 0.237 & 2.58 & 98  & 0.213  & 0.185  & 6.61 & 991 & 0.030 & 0.063  & 8.80  & 1967 & 0.050  & 0.061  & 12.51 & 1979 & 0.080 & 0.083 & 16.11 \\
few-shot (CoT)  & 292 & 0.037 & 0.118 & 3.40 & 100 & 0.070  & 0.050  & 6.62 & 839 & 0.005 & -0.060 & 10.89 & 1850 & 0.230  & 0.284  & 9.04  & 1847 & 0.240 & 0.241 & 10.69 \\ \bottomrule
\end{tabular}
}
\caption{\textbf{Impact of Prompting Strategies on Five STS Datasets} based on \llamatwo (7B), including \medsts, \biosses, \ebmsass, \ustsc, \ustsu under six prompting strategies.}
\label{tab:appendixpromptimpact}
\end{table*}

\section{In-context Learning Case Study}
\label{sec:casestudy}
What are influential factors of the accuracy in in-context learning for STS and NLI?
We first assess the impact of different prompting strategies based on \chatgpt and \llamatwo.

\subsection{Impact of Prompting Strategy}
Using general \stsb and clinical \ncsts test sets, we evaluate the impact of six prompting strategies on STS accuracy, for both \chatgpt and \llamatwo (7B), including  (see \tabref{tab:prompts}): 
\begin{itemize}
    \item Zero-shot 
    \item Zero-shot with annotation guidelines (AG)
    \item Zero-shot with chain of thought (CoT)
    \item Few-shot 
    \item Few-shot with annotation guidelines (AG)
    \item Few-shot with chain of thought (CoT)
\end{itemize}

\paragraph{How to craft prompts?} 
Naive few-shot prompt only shows exemplars to the model, such as five training examples whose similarity score spans from zero to five in our setting. However, the model is often confused about what task it should perform and fail to predict a score.
Thus, we append a task description (same as zero-shot prompt) at the end of demonstrations.
Compared to appending the description at the beginning of the prompt, first showing examples and then elaborating instructions before inputting test cases is easier for model to follow the instruction, resulting in more valid predictions and better accuracy.

For a few-shot prompt with annotation guidelines (see \secref{sec:appendix-prompt}), three components are included: demonstrations, annotation instructions and the task description.
Prompting by the order of task description, instruction and demonstrations, the majority of responses are invalid (441 among the first 500 examples in STS-B), returning ``the score for the given sentence pair is not provided''.
While prompting by first instruction, demonstrations and then the task description, the model will return similarity scores.

Few-shot prompting with chain of thought is crafted with the task description followed by five demonstration examples with an explanation for each one.

\paragraph{How to parse labels from responses?}
One challenge is how to accurately parse the model prediction from a long free-form generation. Many predicted labels do not appear at the beginning, the end or the position requested by the instruction, since the model does not always follow the instruction, particularly for \llamatwo.

For responses of \chatgpt, we use rules and regular expressions to match and parse labels. 
It is hard to parse \llamatwo responses by rules because the irregular positions of the labels, especially responses using \chainofthought. 
To solve this problem, we resort to \llamatwo itself to parse the label out, and then apply simple rules to normalise the results. 
This method alleviates the manual workload to summarise parsing rules, but at the risk of hallucinating inconsistent labels. We observed that \llamatwo would omit decimal places, like parsing similarity score 4.5 to 4, and sometimes generate a new scalar 1.0 without reference in minority cases. 

\subsubsection{\chatgpt}
\paragraph{Zero-shot prompt gives the best correlation based on \chatgpt.}
Results over both general-purpose and clinical STS in \tabref{tab:promptimpact} show that providing annotation guidelines, using chain of thought, and demonstrating labelled examples to the model hurt the STS performance, particularly zero-shot with chain of thought (estimations collapse).
This is counter-intuitive and inconsistent with previous findings that chain of thought and few shots improve the accuracy of reasoning tasks, although \citet{reynolds2021promptprogramming} also showed that cleverly-constructed prompts in a zero-shot setting could outperform prompts in a few-shot setting, implying that, for some tasks, models can achieve better performance by leveraging their existing knowledge than from attempting to learn the task from in-context exemplars.


\begin{table}[ht]
\centering
\resizebox{\columnwidth}{!}{
    \begin{tabular}{c p{7cm}}
        \toprule
        \textbf{No.} & \textbf{Example} \\
        \midrule
        1 
        & S1: A woman is dancing in the rain.  \\
        & S2: A woman dances in the rain outside. \\
        & Label: 5.0  \\
        & Pred: 2.5  \\
        \midrule
        2
        & S1: A man is playing the guitar and singing.  \\
        & S2: A man sings with a guitar.  \\
        & Label: 4.75  \\
        & Pred: 3.0  \\
        \midrule
        3
        & S1: A man is cutting a pipe with scissors. \\	
        & S2: A man is cutting carpet with a knife.	\\
        & Label: 1.2 \\
        & Pred: 3.0 \\
        \bottomrule
    \end{tabular}
    }
    \caption{Incorrectly predicted examples from the STS-B dataset when using zero-shot prompting with annotation guidelines.}
    \label{tab:zeroshot-ag-errors}
\end{table}

\paragraph{Brief annotation guideline and limited exemplars may mislead models.}
With annotation guidelines, it becomes clear how to label sentence pairs that are completely dissimilar or equivalent, but it also brings ambiguous and subjective distinguishment between what is important information and what are unimportant details (score 2-4).

For examples 1 and 2 in \tabref{tab:zeroshot-ag-errors}, the model explains that \textit{two sentences are expressing the same action (dancing in the rain and singing with guitar) and the highly-similar semantic meaning. However, there is a slight difference in the details mentioned, the similarity score between S1 and S2 can be determined as 2.5 and 3.0}.
This suggests that the model fully understands the meaning of two sentences, but fails to assign a correct similarity score.

Similar for No.3, \chatgpt analyses that there are differences in important details between S1 and S2: \textit{pipe} vs. \textit{carpet} and \textit{scissors} vs. \textit{knife}, but it assigns the similarity score of 3.0.
We find for most cases, the reasoning steps are entirely correct, but the model tend to assign a score around 3.0, either two sentences differ significantly in key points or slightly on details.
The model is puzzled by \textit{detail/important information} in guidelines and loses rational judgement.

\paragraph{Why does Zero-shot \chainofthought collapse?}
The rationale behind \chainofthought is improving the performance of reasoning tasks by allowing generative model to infer step by step, instead of outputting results directly.
In the context of STS, reasoning could be either calculating a similarity score quantitatively step by step, or explaining why.

By prompting \chatgpt using zero-shot \chainofthought, it is found to give detailed steps of how to calculate a similarity score using different metrics and features (e.g., tokenise, stem, obtain IF-IDF and calculate cosine similarity). 
Many responses analyse similarity score on axes of sentence structure, bag of words, topics and other aspects between two sentences.

Generally, these scores will be summed up and re-scaled to 0-1 or to 0-5, and sometimes they will be cut by the maximum range of 5 without considering the meaning behind the score. 
Such casual and inconsistent re-scaling creates a situation where the predictions are evaluated in different scales.
Sometimes, these scores conflict with each other --- some are low and some are high, and the model will respond that it is difficult to determine the final score.

Coarse measurements highlight that some specific aspects, such as lexicon overlap and sentence structure, overlook the comparison of the overall semantics.
Moreover, careless re-scaling neglects the meaning behind the score, and the combination substantially hurts the accuracy for STS.
Thus, we guide the model to provide explanations in a few-shot \chainofthought.

\subsubsection{\llamatwo}
\begin{figure*}[t!]
	\centering
	\includegraphics[scale=0.6]{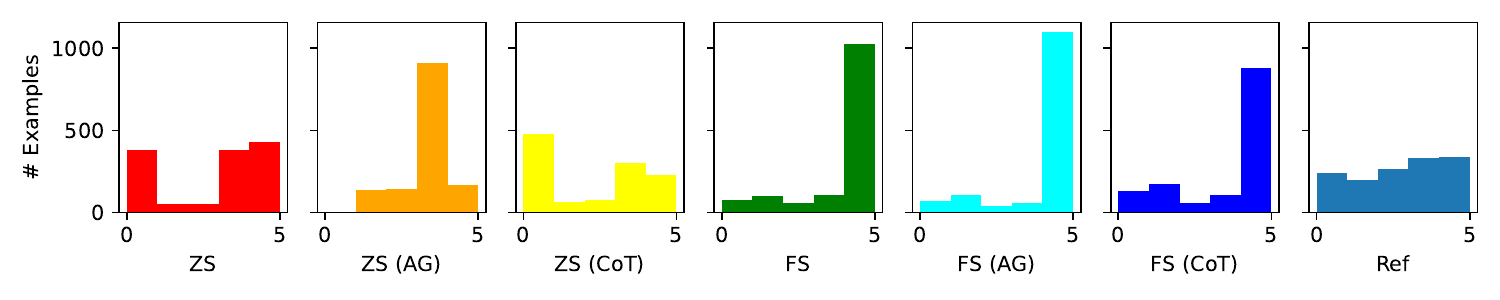} \\
        \includegraphics[scale=0.6]{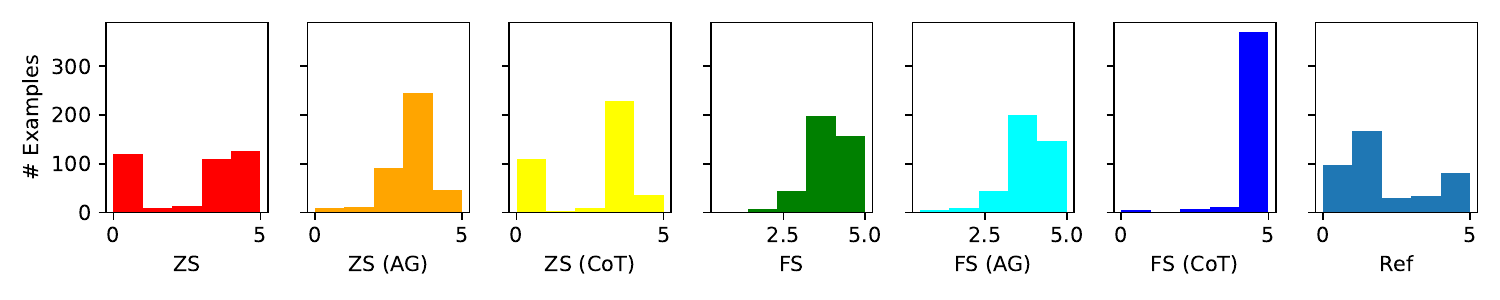} \\
	\caption{Similarity Score distribution of \stsb (top) and \ncsts (bottom) by \llamatwo (7B). Ref=Gold labels}
	\label{fig:score-dist-lllama2}
\end{figure*}

We can further observe that \llamatwo (7B) shows extremely poor performance for both \stsb and \ncsts, particularly with zero-shot prompts: $r$<0.15 (w/wt \chainofthought). 
Using a few-shot (\chainofthought) prompt yields the best correlation $r$=0.67 for \stsb, and the few-shot prompting result for \ncsts is $r$=0.33. 
The results for the other five STS datasets we experimented with also show very low correlations, and few-shot prompting (with/without \chainofthought) yields the best accuracy (see \tabref{tab:appendixpromptimpact}). 
Reflected as the distribution in \figref{fig:score-dist-lllama2}, the predicted score distributions for all prompts deviate significantly from the gold label distribution.
\llamatwo using three few-shot prompts tends to predict scores close to 5.

We find that the low accuracy on the one hand results from the failure of STS modelling of \llamatwo, on the other hand, is partially attributed to the imprecise parsing.
That is, not all predicted labels can be accurately parsed from the generated responses by automatic strategies.
We pass the hard-parsed cases, so the number of valid labels is less than the size of the full test set.
Considering the number of valid cases and the performance, we use few-shot without guidelines and \chainofthought on STS, in the following experiments of \llamatwo.

\paragraph{Impact of Parsing Strategies:}
We find that responses by few-shot prompt is easier to parse by rules.
\tabref{tab:impact-parse-strategy} compares Pearson correlation of predictions parsed by rules and \llamatwo.
Overall, rule-based parsing empirically performs better than parsing by \llamatwo itself on few-shot responses.
Accuracy of \llamatwo (13B) is slightly impacted by parsing strategies, while \llamatwo (7B) is influenced significantly.
We speculate that larger LLMs not only can more accurately parse labels, they are also more capable to follow instructions and generate easily-parsed responses.
\begin{table}[!t]
\centering
\resizebox{\columnwidth}{!}{
    \begin{tabular}{l c c c c c c c}
        \toprule
        Dataset & \stsb & \biosses & \ebmsass & \medsts & \ncsts & \ustsc & \ustsu \\
        \midrule
        \multicolumn{3}{l}{\textbf{\llamatwo (7B)} }\\
        Rules &     0.528 & 0.181 & 0.078 & 0.278 & 0.328 & 0.038 & 0.076 \\
        \llamatwo & 0.506 & 0.151 & 0.081 & 0.255 & 0.327 & 0.033 & 0.076 \\
        \midrule
        \multicolumn{3}{l}{\textbf{\llamatwo (13B)} }\\
        Rules &     0.584 & 0.254 & 0.189 & 0.186 & 0.254 & 0.004 & 0.107 \\
        \llamatwo & 0.583 & 0.255 & 0.195 & 0.186 & 0.252 & 0.003 & 0.11 \\
        \bottomrule
    \end{tabular}
    }
    \caption{\textbf{Impact of parsing strategy:} Pearson correlation ($r$) of seven STS datasets based on few-shot prompt under \llamatwo 7B (top) and 13B (bottom). Rule-based parsing overall performs better than parsing by \llamatwo itself on responses by few-shot prompt. Accuracy of \llamatwo (13B) is slightly impacted by parsing strategies.}
    \label{tab:impact-parse-strategy}
\end{table}

\subsubsection{Zero-shot vs. Few-shot for NLI}
Given that there isn't complex annotation guidelines for NLI, and \chainofthought is demonstrated less improvements, we only compare the naive zero-shot and few-shot prompts for NLI.
\tabref{tab:nli-prompt-impact} shows that for both \llamatwo 7B and 13B, few-shot prompt can achieve either higher or comparable F1-score than zero-shot prompt across three NLI datasets.
This is consistent with the STS task using \llamatwo.
Therefore, on \chatgpt, we follow STS to use zero-shot prompt for NLI as well.
\begin{table}[!t]
\centering
\resizebox{\columnwidth}{!}{
    \begin{tabular}{l | c c c | c c c}
        \toprule
        Model$\rightarrow$ & \multicolumn{3}{c|}{\textbf{\llamatwo (7B)}} & \multicolumn{3}{c}{\textbf{\llamatwo (13B)}} \\
        Dataset$\rightarrow$ & S & M & MED & S & M &  MED \\
        \midrule
        Few-shot &  \textbf{0.375} & \textbf{0.306} & \textbf{0.312} & \textbf{0.319} & 0.321 & \textbf{0.414}\\
        Zero-shot & 0.204 & 0.288 & 0.253 & 0.205 & \textbf{0.323} & 0.293\\
        \bottomrule
    \end{tabular}
    }
    \caption{\textbf{F1-score by Zero vs. Few-shot for NLI} over \chaossnli (S), \chaosmnli (M) and \mednli (MED) under \llamatwo 7B and 13B.}
    \label{tab:nli-prompt-impact}
\end{table}

\begin{figure}[t!]
	\centering
        \includegraphics[scale=0.45]{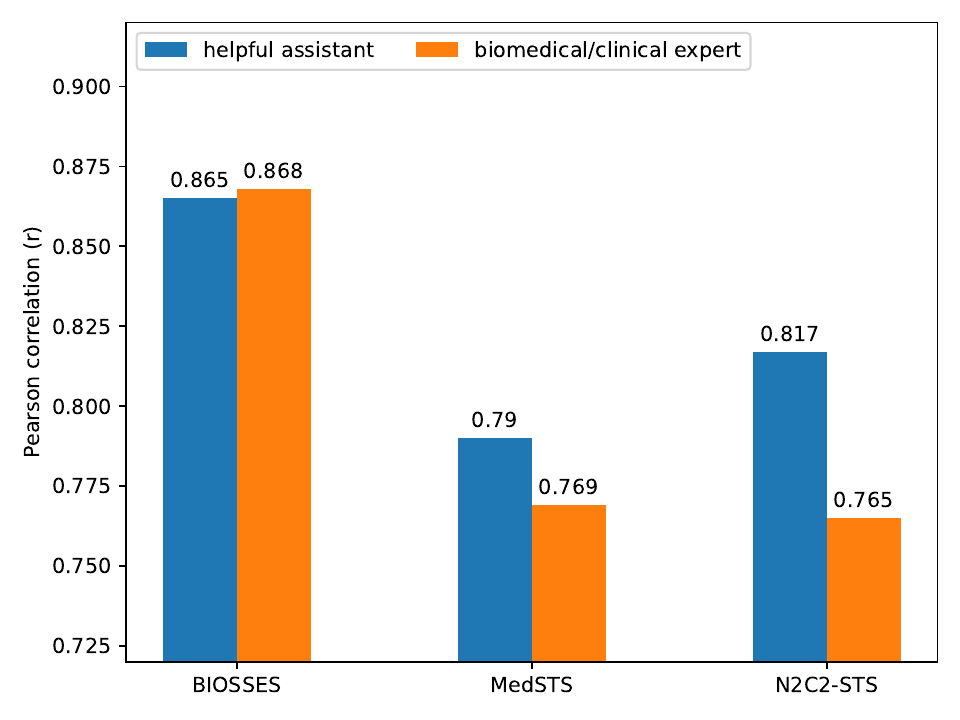}
	\caption{The impact of system role on the performance of domain datasets using \chatgpt.}
	\label{fig:systemrole-impact}
\end{figure}
\subsection{Impact of Metadata in Prompt}
Will setting the system role as domain expert result in better performance in domain datasets?
Do Chinese prompts perform better than English prompt on Chinese datasets?
We try to answer the two questions in this section.

\paragraph{System role and context}
On the biomedical STS dataset \biosses and two clinical datasets (\medsts and \ncsts), we compare the correlation with system role (pre-context) set as ``helpful assistant'' vs. ``biomedical/clinical expert''.
\figref{fig:systemrole-impact} shows that the accuracy either declines or is the same when setting the system role to domain expert from general assistant. 
Similarly, changing zero-shot prompt to ``determine the similarity between the following two sentences (S1, S2) \textit{in the biomedical context with domain knowledge}'' does not help either. 
Combining them yields BIOSSTS correlation declining from 0.868 to 0.848.

\begin{table}[!t]
\centering
\resizebox{\columnwidth}{!}{
    \begin{tabular}{l l l l l}
        \toprule
        Dataset & lan\_instruction & $r\uparrow$ & $\rho \uparrow$ & MSE $\downarrow$ \\
        \midrule
        \ustsc (high) & English & \textbf{0.556} & \textbf{0.551} & \textbf{2.97} \\
        \ustsc (high) & Chinese & 0.461 & 0.503 & 5.00 \\
        \ustsu (low) & English & \textbf{0.552} & \textbf{0.465} & \textbf{3.09} \\
        \ustsu (low) & Chinese & 0.472 & 0.435 & 5.42 \\
        \bottomrule
    \end{tabular}
    }
    \caption{Correlation ($r$, $\rho$) and MSE on Chinese \ustsc (high human disagreement in labelling) and \ustsu (low human disagreement) test sets using \chatgpt (helpful assistant), by \textit{en} and \textit{zh} prompts.}
    \label{tab:usts}
\end{table}

\newpage
\paragraph{Language of the prompt}
Evaluating LLMs on non-English benchmarks, we have two choices for the language of the prompt: English prompt that the LLM has seen more than other languages in training and tuning, and corresponding language instruction that is consistent with the input content.

Based on a Chinese STS corpus \usts with two subsets: \ustsc with high human disagreement in labelling and \ustsu with low human disagreement, we compare the results using English vs. Chinese zero-shot prompts in \tabref{tab:usts}.
Using English instruction shows higher correlation and smaller MSE than using Chinese instruction.
For both subsets, correlations between the predicted score and the gold label by averaging annotations of all raters are both extremely low (around 0.5), and MSE is large.
This implies that it is challenging for \chatgpt to correctly estimate semantic similarity scores for Chinese sentence pairs in \usts, regardless of high or low human disagreement.

Moreover, for fine-tuned STS models based on BERT or cosine similarity based on semantic representation of two sentences, it is easier to predict the average score for \ustsu than \ustsc.
\chatgpt does not seem to perceive the degree of human disagreement in labelling, showing higher accuracy on more uncertain subset \ustsc.

\section{Prompting Strategies}
\label{sec:appendix-prompt}
GPT-3~\citep{brown2020gpt3} demonstrated that LLMs are strong few-shot learners, where fast in-context learning can be achieved through prompting strategies. Through a handful of demonstration examples encoded as prompt text in the input context, LLMs are able to generalise to new examples and new tasks without any gradient updates or fine-tuning. 
The remarkable success of in-context few-shot learning has spurred the development of many prompting strategies including scratchpad, chain-of-thought, and least-to-most prompting, especially for multi-step computation and reasoning problems such as mathematical problems.
In this study for STS and NLI, we focus on standard zero-shot, few-shot, chain-of-thought, and self-consistency prompting as discussed below.

\textbf{Few-shot:} 
The standard few-shot prompting strategy was introduced with GPT-3. 
The prompt to the model is designed to include few-shot examples describing the task through text-based demonstrations. These demonstrations are typically encoded as input–output pairs. 

After the prompt, the model is provided with an input and asked to generate a prediction. 
We identify five demonstration input–output examples for each dataset and we craft the few-shot prompts. 


\textbf{Zero-shot:} 
The zero-shot prompting typically only involves an instruction describing the task without any examples (see \tabref{tab:prompts}).

\textbf{Chain of thought (\chainofthought) and Explanation:}
\chainofthought~\citep{wei2022chain} involves augmenting each few-shot example in the prompt with a step-by-step breakdown and a coherent set of intermediate reasoning steps towards the final answer. 

This approach is designed to mimic the human thought process when solving problems that require multi-step computation and reasoning. 
\chainofthought prompting can elicit reasoning abilities in sufficiently powerful LLMs and can dramatically improve the performance for certain tasks, e.g., when solving mathematical problems. 

A variant of \chainofthought is to prompt LLMs with explanation, instead of label-only prediction. It shows to be more robust over hard and adversarial NLI examples, since it forces models to conduct rationalise-then-predict~\citep{kavumba-etal-2023-prompting}. 
That is to learn what NLI task intended to learn, rather than superficial cues, such as association between label \textit{contradict} and token \textit{not} in hypothesis (models are ``right for the wrong reason'').

This is consistent with the finding presented by \citet{zhang2023language}, LLMs indeed have the knowledge/capability to answer questions correctly if we prompt it to rationalise step by step, instead of asking them to give a \textit{Yes/No} answer in the first token, where they tend to predict wrongly.
Multiple steps or explanation prompting may allow models to ``think over'' and then infer answers, decreasing the error rate resulting from \textit{quick quiz} (less time to think).

Overall, these findings indicate that prompting large language models by multi-step reasoning or giving explanations before predicting labels can lead to robust performance over hard and adversarial answers.
On top of these findings, when proposing prompts, we allow models to generate explanation by ``thinking'' multiple steps before predicting the final label, to fully unlock LLM's capabilities.

\textbf{Self-consistency}
A straightforward strategy to improve the performance of a model on the multiple-choice benchmarks is to prompt and to sample multiple decoding outputs from the model. The final answer then is the one that received the majority vote. 
This idea was introduced as self-consistency. 
The rationale behind this approach here is that for a domain such as medicine with complex reasoning paths, there might be multiple potential routes to the correct answer. 
Marginalising out the reasoning paths can lead to the most consistent answer. The self-consistency prompting strategy led to particularly strong improvements in reasoning tasks, and we adopted the same approach for our datasets.

\paragraph{Annotation Guidelines}
The instruction: 0 denotes complete dissimilarity between two sentences; 1 shows that two sentences are not equivalent but are topically related to each other while score of 2 indicates that two sentences agree on some details mentioned in them. 3 implies that there are some differences in important details described in two sentences while a score of 4 represents that the differing details are not important. And 5 represents that two sentences are completely similar. 

\section{White-box Label-token Probability}
\label{sec:whiteboxlabelprob}
\begin{table}[ht!]
\centering
\resizebox{\columnwidth}{!}{
    \begin{tabular}{l | c c c | c c c }
        \toprule
        Model$\rightarrow$ & \multicolumn{3}{c|}{\textbf{\llamatwo (7B)}} & \multicolumn{3}{c}{\textbf{\llamatwo (13B)}} \\
        Dataset$\downarrow$
        & T1\_is$\uparrow$ & T1\_prob$\uparrow$ & T3\_has$\uparrow$ 
        & T1\_is$\uparrow$ & T1\_prob$\uparrow$ & T3\_has$\uparrow$ \\
        \midrule
        \medsts & 100.0 & 0.818 & 100.0 & 100.0 & 0.754 & 100.0 \\
        \biosses & 100.0 & 0.840 & 100.0 & 100.0 & 0.723 & 100.0\\
        \ustsc & 100.0 & 0.751 & 100.0 & 100.0 & 0.664 & 100.0 \\
        \midrule
        \mednli & 99.9 & 0.868 & 100.0 & 96.3 & 0.797 & 98.6\\
        \chaosnli & 98.0 & 0.795 & 99.0 & 85.0 & 0.752 & 93.0\\
        \bottomrule
    \end{tabular}
    }
    \caption{\textbf{Can the first token be in the label space:} T1\_is = the percentage of examples where top1 (highest probability) token is in the label space, T1\_prob = the average probability of the top1 probability if it is in the label space, T3\_has = the percentage of examples where top3 tokens contain label-space tokens.}
    \label{tab:topprobtokens}
\end{table}

\section{\secref{sec:humanopinion} Supplementary Information}
\paragraph{Ten runs under the same role} in \tabref{tab:tenruns}.
\begin{table}[ht!]
\centering
\resizebox{\columnwidth}{!}{
    \begin{tabular}{c | c c c c | c c c}
        \toprule
        Dataset$\rightarrow$  & \multicolumn{4}{c|}{\chaosnli} & \multicolumn{3}{c}{\ustsc} \\
        \textbf{Run No.} $\downarrow$ & Acc$\uparrow$ & Prec$\uparrow$ & Recall$\uparrow$ & F1-macro$\uparrow$ & $r\uparrow$ & $\rho \uparrow$ & MSE $\downarrow$ \\
        \midrule
        1 & 0.555 & 0.532 & 0.526 & 0.522 & 0.758 & 0.778 & 2.77 \\
        2 & 0.500 & 0.476 & 0.470 & 0.467 & 0.675 & 0.746 & 3.27 \\
        3 & 0.530 & 0.502 & 0.500 & 0.497 & 0.699 & 0.741 & 3.02 \\
        4 & 0.530 & 0.509 & 0.519 & 0.510 & 0.666 & 0.695 & 3.13 \\
        5 & 0.510 & 0.496 & 0.466 & 0.467 & 0.707 & 0.715 & 2.96\\
        6 & 0.540 & 0.528 & 0.526 & 0.518 & 0.702 & 0.749 & 3.15\\
        7 & 0.520 & 0.494 & 0.492 & 0.488 & 0.718 & 0.765 & 3.00\\
        8 & 0.560 & 0.547 & 0.553 & 0.538 & 0.675 & 0.719 & 3.19\\
        9 & 0.555 & 0.527 & 0.527 & 0.523 & 0.721 & 0.749 & 2.91 \\
        10 & 0.565 & 0.540 & 0.533 & 0.530 & 0.707 & 0.736 & 2.90 \\
        \midrule
        Ensemble & 0.570 & 0.547 & 0.544 & 0.541 & 0.809 & 0.840 & 2.79 \\
        \bottomrule
    \end{tabular}
    }
    \caption{Ten runs for \chaosnli under the role of NLP PhD student and \ustsc under a linguistic expert. Ensemble refers to majority voting for NLI and averaging for STS over ten runs.}
    \label{tab:tenruns}
\end{table}

\begin{figure}[ht!]
	\centering
        \includegraphics[scale=0.38]{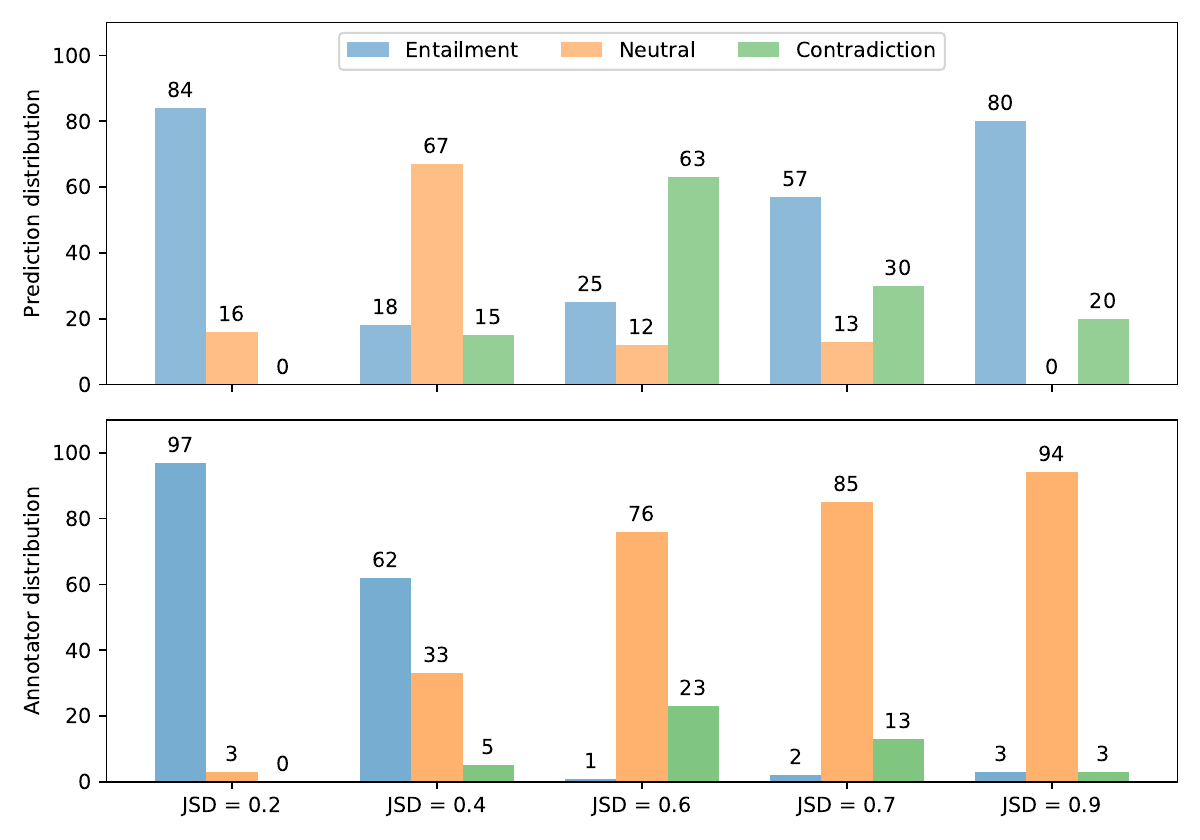}
	\caption{\chaosnli five examples. JSD between distribution of annotators and \chatgpt distributions ranges from 0.2, 0.4, 0.6, 0.7 to 0.9.}
	\label{fig:chaosnli-samples}
\end{figure}
\paragraph{What does JSD=0.2 mean if reflected to NLI labels?}
JSD is symmetric and ranged from 0.0 to 1.0.
Reflected to a specific label, how large differences between two distributions will result in JSD=0.2?
We randomly selected an example whose JSD between annotators and ten roles equal to 0.2, 0.4, 0.6, 0.7, and 0.9, shown on \figref{fig:chaosnli-samples}.

We can see that when JSD$\le$0.2, the majority label always remain the same, while it changes to another when JSD is greater than 0.2.

\paragraph{Ten system roles}
\begin{itemize}
    \item You are a helpful assistant
    \item You are a helpful assistant, doing well in semantic reasoning and identifying sentence pair relationship
    \item You are a helpful assistant, good at doing natural language inference task
    \item You are an expert in natural language processing
    \item You are a PhD student in natural language processing
    \item You are a data annotator
    \item You are a linguistic expert
    \item You are a Google senior engineer
    \item You are a professional data scientist
    \item You are a five-year old child
\end{itemize}

\end{document}